\documentclass[letterpaper, 10 pt, conference]{ieeeconf}
\IEEEoverridecommandlockouts
\usepackage{array}
\usepackage{textcomp}
\usepackage{stfloats}
\usepackage{url}
\usepackage{verbatim}
\usepackage{graphicx}
\usepackage{subcaption}
\usepackage{cite}
\usepackage{svg}
\usepackage{amsmath}
\usepackage{amssymb}
\usepackage{amsfonts}
\usepackage{multirow}
\usepackage{float}
\usepackage{tabularx,ragged2e}
\usepackage{booktabs}
\usepackage{adjustbox}

\usepackage{xcolor}
\usepackage[linesnumbered, ruled, vlined]{algorithm2e}

\SetCommentSty{mycommfont}
\usepackage[colorlinks,citecolor=red,urlcolor=blue,bookmarks=false,hypertexnames=true,linkcolor=violet]{hyperref}
\graphicspath{ {./images/} }
\usepackage{cite}
\usepackage{fancyhdr}

\title{\LARGE \bf Learning Whole-body Manipulation for Quadrupedal Robot}

\author{Seunghun Jeon$^{1\dagger}$, Moonkyu Jung$^{1\dagger}$, Suyoung Choi$^{1\dagger}$, Beomjoon Kim$^{*2\dagger}$ and Jemin Hwangbo$^{*1\dagger}$% <-this % stops a space
\thanks{This work was supported in part by Korea Evaluation Institute of Industrial Technology (KEIT) funded by the Korea Government (MOTIE) under Grant No.20018216.}
\thanks{$\dagger$ Korea Advanced Institute of Science and Technology (KAIST)}
\thanks{$*$ corresponding author}
\thanks{$^{1}$ Robot Artificial Intelligence Laboratory  $^{2}$Intelligent Mobile-Manipulation  Laboratory}
\thanks{{\tt\small jhwangbo@kaist.ac.kr, beomjoon.kim@kaist.ac.kr}}
}

\begin{document}
\maketitle
\thispagestyle{empty}
\pagestyle{empty}

\thispagestyle{fancy}
\fancyhf{}
\renewcommand{\headrulewidth}{0pt}
\cfoot{This work has been submitted to the IEEE for possible publication. Copyright may be transferred without notice, after which this version may no longer be accessible.}

\begin{abstract}
    We propose a learning-based system for enabling quadrupedal robots to manipulate large, heavy objects using their whole body. Our system is based on a hierarchical control strategy that uses the deep latent variable embedding which captures manipulation-relevant information from interactions, proprioception, and action history, allowing the robot to implicitly understand object properties. We evaluate our framework in both simulation and real-world scenarios. In the simulation, it achieves a success rate of 93.6 $\%$ in accurately re-positioning and re-orienting various objects within a tolerance of 0.03 $m$ and 5 $^\circ$. Real-world experiments demonstrate the successful manipulation of objects such as a 19.2 $kg$ water-filled drum and a 15.3 $kg$ plastic box filled with heavy objects while the robot weighs 27 $kg$. Unlike previous works that focus on manipulating small and light objects using prehensile manipulation, our framework illustrates the possibility of using quadrupeds for manipulating large and heavy objects that are ungraspable with the robot's entire body. Our method does not require explicit object modeling and offers significant computational efficiency compared to optimization-based methods. The video can be found at \href{https://youtu.be/fO_PVr27QxU}{https://youtu.be/fO\_PVr27QxU}.
\end{abstract}


\section{Introduction}
% \IEEEPARstart{W}{hile}

\begin{figure*}[!ht]
    \centering
    \includegraphics[width=\textwidth]{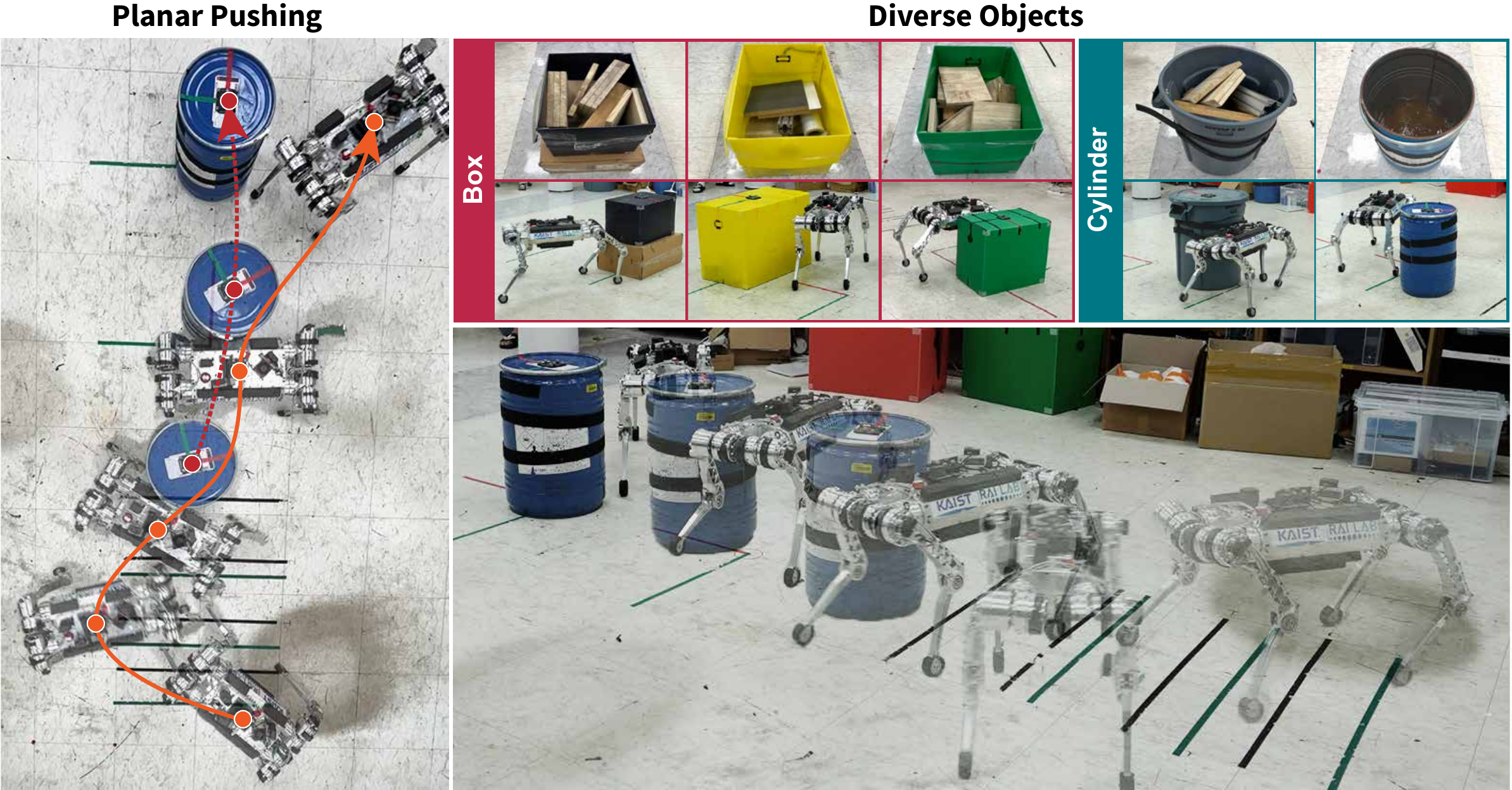}
    \caption{\textbf{Whole-body manipulation for planar pushing}: The proposed system enables a quadrupedal robot, weighing 27 $kg$, to manipulate its entire body to re-pose an object in a 2D plane. Our system does not require the object's physical properties or dimensions but achieves an implicit understanding of the object's properties through real-time interactions. Experiments were conducted on a total of five different objects with diverse properties. The robot can adapt even when the object's physical properties change dynamically, such as when dealing with a 19.2 $kg$ water-filled drum that is approximately 70 $\%$ of the robot's weight.}
    \label{fig: Whole-body manipulation for planar pushing}
\end{figure*}

While recent quadrupedal robots have a remarkable capability to maneuver through a diverse set of challenging and complex terrains \cite{lee2020learning, miki2022learning, choi2023learning}, they still lack the ability to handle tasks requiring interaction with objects and the environment. There have been attempts to attach manipulators to quadrupedal robots \cite{xin2022loco, bellicoso2019alma, ma2022combining, sleiman2021unified}; however, in industrial environments, quadrupeds often face situations where they need to move objects that are large, heavy, and have agnostic physical properties (e.g. shape, mass, inertia, COM). Consequently, prehensile manipulation in such an environment becomes challenging. Our goal is to endow quadrupeds with the capability to manipulate such objects from high-dimensional sensory data using their entire body. This is a challenging task that involves not only determining torque commands to balance and locomote amidst contact with the object but also reasoning about the physical properties of objects to determine a long sequence of actions that would move the object to the target location. 
% Our goal is to endow quadrupeds with such capability from high-dimensional sensory data.

The conventional approach to whole-body manipulation involves designing an optimization-based planner and controller by carefully modeling the dynamics of the object, environment, and robot \cite{rigo2022contact, sombolestan2022hierarchical, ferrolho2022roloma, mittal2022articulated, bellicoso2019alma}. However, this approach has two drawbacks. Firstly, it incurs a significant computational cost due to hybrid dynamics and decision variables involving binary contact sequences and continuous motions under several dynamics constraints that change with respect to contact decisions. With simplifying assumptions, we can significantly reduce the computational cost, but they come at the expense of limited robot motion types \cite{rigo2022contact, sombolestan2022hierarchical}. Secondly, these models rely on the accurate estimation of explicit object information, such as its shape and physical parameters, from sensory data, and the action history itself is a challenging task.
% which is a challenging task when relying at proprioception and action history.

Recently, learning-based methods for whole-body manipulation have been proposed \cite{shi2021circus, ji2023dribblebot, fu2023Deep, kumar2022cascaded, cheng2023legs}. Model-free approaches offer advantages over conventional optimization-based methods by enabling robots to adapt to diverse environments using the generalization capability of neural network (NN) \cite{ji2023dribblebot, kumar2022cascaded}. Additionally, the computation of the actions comes down to making predictions from a NN, instead of a complex optimization procedure, significantly reducing the computational cost. However, previous works have primarily focused on training systems to manipulate a single object, such as a ball \cite{ji2023dribblebot} or a cylinder \cite{kumar2022cascaded}, which introduces inherent challenges when dealing with objects of various types, geometries, and physical properties. The difficulty arises from the need to adjust forces depending on the specific object type.

In this paper, we tackle the problem with a hierarchical control strategy. The high-level controller outputs velocity commands based on the state of the robot and the object, while the low-level controller outputs desired joint position that leads joint torque to track given commands. Additionally, we show that the difficulties mentioned above ---manipulating a wide range of objects with diverse physical properties--- can be mitigated by using a deep latent variable encoding model that embeds object information from multiple interactions, proprioception, and action history into a low-dimensional latent vector. This enables an implicit understanding of object physical properties, akin to how humans manipulate an object when they lack knowledge about the object physical properties. Initially, humans rely on prior knowledge to determine the force's direction and magnitude. Subsequently, through physical interactions, they refine their understanding, considering factors like the applied force vector and perceived object motion. 

We conduct extensive evaluations, achieving a success rate of 93.6 $\%$ in re-positioning various objects to the desired pose within 0.03 $m$ and 5$^\circ$ in simulation. Moreover, we successfully deployed the learning-based system, trained in simulation, directly to the real-world scenario without any fine-tuning or additional training. The results in the real world, show that our proposed system can manipulate heavy and complex objects without a single failure, such as water-filled drums weighing 19.2 $kg$ and plastic boxes filled with wooden furniture weighing 15.3 $kg$, as shown in Figure. \ref{fig: Whole-body manipulation for planar pushing}.
\section{Related Work}

\subsection{Whole-body manipulation for legged robots}
Rigo et al. \cite{rigo2022contact} propose a real-time controller for non-prehensile loco-manipulation that tracks a given object trajectory under the given dynamics and assumes canonical push dynamics. The proposed method could re-position a 5 $kg$ box-shaped object to the desired pose by optimizing contact point location and force magnitude and consequently derive desired joint torques via a hierarchical Model Predictive Control (MPC) framework. Sombolestan et al. \cite{sombolestan2022hierarchical} present an approach for whole-body manipulation tasks on unknown objects and terrains without prior knowledge. They propose a real-time adaption method that can push an object where physical properties vary with time weighing up to 7 $kg$, even on slopes in a single direction. These works demonstrate the potential for optimization-based approaches to enable robots to perform loco-manipulation for pushing. However, these approaches assume a single, flat-surface contact at the pre-defined robot's head, not the whole body of the robot.

Learning-based approaches also have been considered in whole-body manipulation. Ji et al. \cite{ji2023dribblebot} successfully overcome challenges related to perceiving and controlling a ball on different terrains. The robot action policy is trained to achieve desired object-centric velocity command under assorted surroundings via Reinforcement Learning (RL). Kumar et al. \cite{kumar2022cascaded} presents a cascaded compositional residual learning approach for complex interactive behaviors. The approach decomposes a complex behavior into a sequence of simpler sub-behaviors and learns a residual policy for each sub-behavior. The residual policies are then combined to form a policy for the complex behavior. They evaluate their approach to a variety of tasks, including cylindrical object pushing. In this task, the policy was able to successfully push the object within 0.1 $m$ of the target location 89 $\%$ of the time. 
These works demonstrate a learning-based approach can be successfully applied to dynamic whole-body manipulation. However, previous studies have not considered generalization over a variety of objects, and they have not considered situations where the physical properties of objects change significantly in real-time, or where the goal includes orientation. We address these aspects in this work.

\subsection{Object system identification through physical interactions}
Various methods have been developed for object system identification through physical interactions, aiming to estimate the object parameters and, consequently, the motion. Wang et al. \cite{wang2020swingbot} propose SwingBot, which utilizes tactile feedback data collected during the execution of predefined motions to implicitly encode physical properties. Kloss et al. \cite{kloss2020accurate} employs the Kalman filter to estimate object physical properties, including the center of mass, friction, and mass. Song et al. \cite{song2020probabilistic} propose a probabilistic model for identifying the mass and friction properties of objects with unknown material properties, which is trained on a simulated system and is effective in real-world experiments. Mavrakis et al. \cite{mavrakis2020estimating} propose a data-driven method for estimating the object's inertial parameters, which is trained on a large dataset of simulated and real-world pushes. Xu et al. \cite{xu2019densephysnet} introduce DensePhysNet, a neural network-based approach to implicitly represent object physical properties. Fragkiadaki et al. \cite{fragkiadaki2015learning} predict ball motion under a pushing force using object-centric prediction approaches. These methods either explicitly estimate physical parameters or implicitly encode them using neural networks, contributing to the understanding and prediction of object motion based on real-world experiences. Our system involves predicting object properties implicitly through a neural network using physical interactions.

\subsection{Privileged learning and adaptation}

Many studies employ a simulation-to-real adaptation approach, where the policy is trained using privileged information (i.e. teacher policy) that is only observable in simulation and remains agnostic to the real-world scenario. This enables the policy to quickly achieve high performance when deployed in the real world. In some studies, the vision sensory data, robot, and environment's physical properties are implicitly estimated \cite{lee2020learning, kumar2021rma, fu2023Deep}. For instance, Lee et al. \cite{lee2020learning} distilled the teacher policy's action output after training through imitation learning. Kumar et al. \cite{kumar2021rma} trained the privileged information encoder and adapted the encoded latent feature space via 2-step training. Fu et al. \cite{fu2023Deep} remove the two-phase teacher-student scheme, and regularize environment extrinsics latent to avoid large deviations. While others explicitly estimated the privileged information, Ji et al. \cite{ji2022concurrent} concurrently trained the state estimator for the base linear velocity, foot height, and contact probability. Additionally, some studies \cite{nahrendra2023dreamwaq, ma2023learning} use an asymmetric structure, where the critic evaluates the value only through privileged information, while the actor makes actual decision-making based on observable values. We leverage privileged learning in an implicit latent space that encodes an object physical properties.
\section{Learning whole-body manipulation for planar pushing} \label{Learning whole-body manipulation for planar pushing}

\begin{figure*}[ht!]
    \centering
    \includegraphics[width=\textwidth]{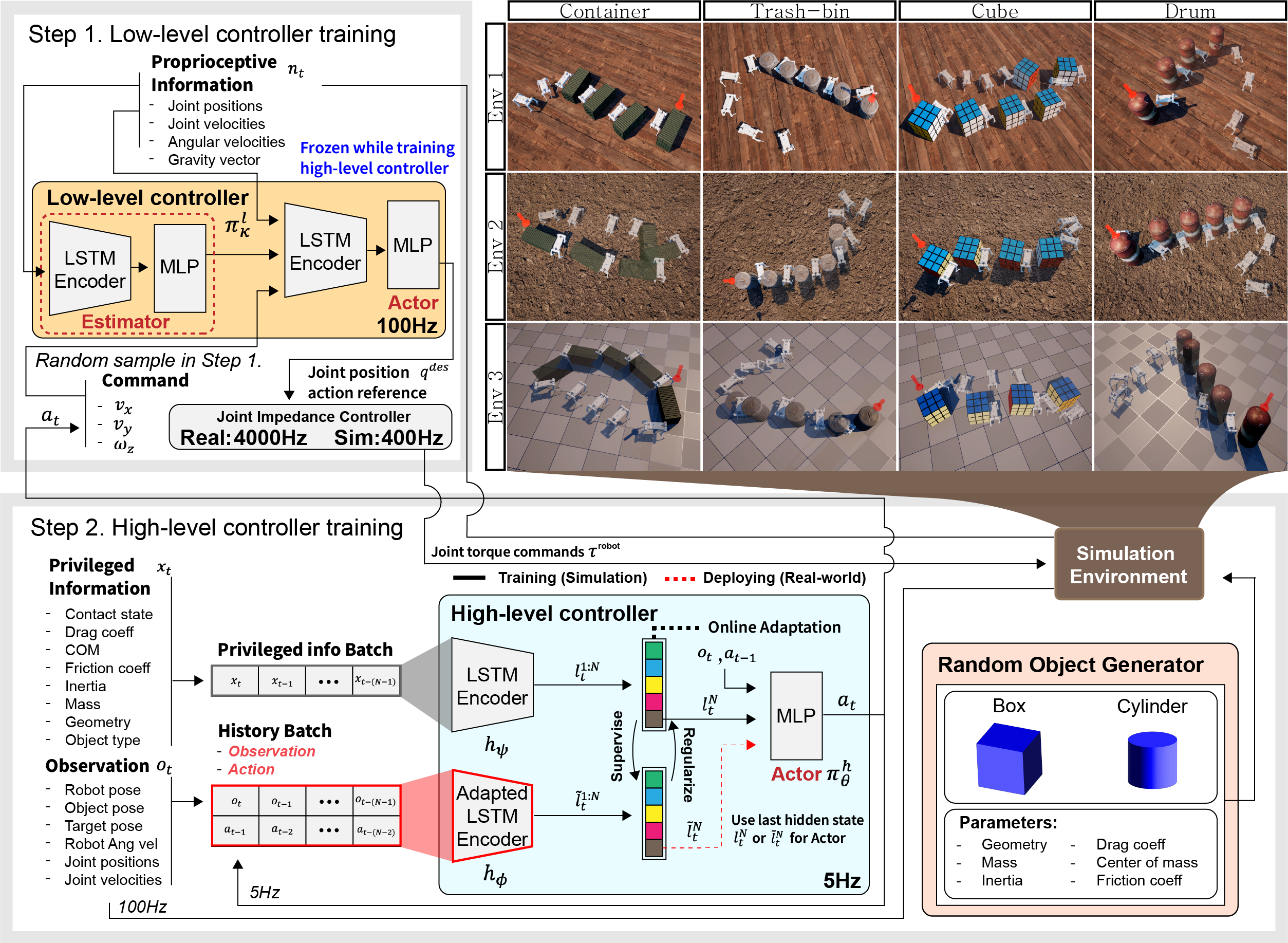}
    \caption{\textbf{Overall framework}: In step 1, we train the low-level controller alone via reinforcement learning. The low-level controller tracks a given velocity command, following the method in \cite{choi2023learning}. In step 2, we freeze the low-level controller and only train the high-level controller. The high-level policy outputs an action command from a current observation $o_t$, previous action $a_{t-1}$, and a latent vector that encodes a history batch of privileged information $x_t$. It is also trained via reinforcement learning. Adaptation is performed online on the encoder of the high-level controller through supervised learning and regularization of latent vectors, following the method in \cite{fu2023Deep}. The high-level controller generates the robot body $se(2)$ velocity every 5Hz, and this action command is fed to the low-level controller. The low-level controller generates the desired joint positions at 100Hz, which are converted to torque via an impedance controller.}
\label{fig: overall flow}
\end{figure*}

Our task is to determine a sequence of robot joint torques that manipulates an arbitrary object to the target pose using the robot's whole body. We assume that we do not have prior knowledge about the physical properties of the object, but have complete knowledge of the object's and robot's poses. We assure, at the start, there is a considerable distance between the object and the robot, so the robot first has to approach the object and push it to the target pose. Our system has a hierarchical structure, as illustrated in Figure \ref{fig: overall flow}. In the following sections, we provide a detailed description of each component of our system.

% In this paper, we are interested in pushing an arbitrary
% object, following a planned trajectory in terms of x and
% y world frame position and heading angle ψ. We assume
% we know all the geometric and inertial characteristics of
% the object, and we have the feedback on its heading angle
% and center of mass position. Due to the limitations of the
% pushing primitive, to move the object to the desired location,
% we have to align its heading angle toward that location.
% Leveraging the position tracking of the quadruped robot,
% we can optimize the contact point between the robot head
% and object to push forward and, at the same time, rotate
% the object to align the heading angle to the desired one.
% Without changing the contact location, we would not be able
% to control the heading angle of the object. The nonlinearity
% of the loco-manipulation problem is solved by splitting it into
% two separate linear parts, the first responsible for determining
% the required manipulation action to be exerted on the object;
% the second responsible for the locomotion under the effect
% of the contact interaction.

\subsection{Training whole-body manipulation in simulation} \label{Training in simulation} 

\begin{table}[ht!]
\caption{\textbf{Notations}: Variable representation}
\centering
\resizebox{\columnwidth}{!} {
\begin{tabular}[t]{|l|c|c|c|c|c|c|}
\hline
\multicolumn{2}{|c|}{\textbf{Notations}} & \textbf{Components} & \textbf{Dim} & \textbf{Components} & \textbf{Dim} & \textbf{Total} \\ \hline
\multicolumn{1}{|c|}{\multirow{17}{*}{$s_t$}}
& \multicolumn{1}{c|}{\multirow{10}{*}{$o_t$}} 
& $\phi_{robot,x}$ & 3 & $v_{robot}$ & 3 & \multirow{10}{*}{33} \\ \cline{3-6}
& & $w_{robot}$ & 3 & $\phi_{robot}^T(p_{left\_foot} - p_{robot})$ & 3 & \\ \cline{3-6}
& & $\phi_{robot}^T(p_{right\_foot} - p_{robot})$ & 3 & $\frac{p_{object}-p_{robot}}{||p_{object}-p_{robot}||_2}$ & 2 & \\ \cline{3-6}
& & $||p_{object}-p_{robot}||_2$ & 1 & $\frac{p_{target}-p_{object}}{||p_{target}-p_{object}||_2}$ & 2 & \\ \cline{3-6}
& & $||p_{target}-p_{object}||_2$ & 1 & $\frac{p_{target}-p_{robot}}{||p_{target}-p_{robot}||_2}$ & 2 & \\ \cline{3-6}
& & $||p_{target}-p_{robot}||_2$ & 1 & ${\phi_{robot}}^Tv_{object}$ & 3 & \\ \cline{3-6}
& & ${\phi_{robot,x}}^T\phi_{object,x}$ & 1 & $||\phi_{robot,x} \times \phi_{object,x}||_2$ & 1 & \\ \cline{3-6}
& & ${\phi_{robot,x}}^T\phi_{target,x}$ & 1 & $||\phi_{robot,x} \times \phi_{target,x}||_2$ & 1 & \\ \cline{3-6}
& & ${\phi_{object,x}}^T\phi_{target,x}$ & 1 & $||\phi_{object,x} \times \phi_{target,x}||_2$ & 1 & \\ \cline{2-7}

& \multicolumn{1}{c|}{\multirow{4}{*}{$x_t$}} 
& $Object \ type$ & 3 & $Dimension$ & 3 & \multirow{4}{*}{22} \\ \cline{3-6}
& & $m_{object}$ & 1 & $COM_{object}$ & 3 & \\ \cline{3-6}
& & $I_{object}$ & 9 & $Friction \ coeff$ & 1 & \\ \cline{3-6}
& & $Drag \ coeff$ & 1 & $Contact$ & 1 & \\ \cline{2-7}

& \multicolumn{1}{c|}{\multirow{3}{*}{$n_t$}} 
& $\phi_{robot,z}$ & 3 & $w_{robot}$ & 3 & \multirow{4}{*}{33} \\ \cline{3-6}
& & $v_{robot}$ & 3 & $q$ & 12 & \\ \cline{3-6}
& & $\dot{q}$ & 12 &  &  & \\ \hline

\end{tabular}
\label{table:notations}
}
\end{table}

We first define the variables as shown in Table \ref{table:notations} for the following descriptions. The state $s_t$ is the concatenation of observation $o_t$, privileged information $x_t$, and robot state $n_t$. Here, $\phi_{\langle \cdot \rangle}$ represents the rotation matrix, and $\phi_{\langle \cdot \rangle, x}$ represents the x-axis component of the rotation matrix (i.e. first column vector). $v_{\langle \cdot \rangle}$ represents the COM velocity, while $w_{\langle \cdot \rangle}$ represent the COM angular velocity. The symbol $\times$ corresponds to the cross product. The variable $p_{\langle \cdot \rangle}$ represents the geometric center position. The object type is represented as a one-hot vector, and the dimension information is represented as $\{$width, depth, height$\}$ of the object. $m_{\langle \cdot \rangle}$ represents the mass, and $I_{\langle \cdot \rangle}$ represents the inertia. $Contact$ represents robot actually contacts the object or not. Additionally, $q$ represents the robot joint position, and $\dot{q}$ represents the robot joint velocity. \\

\paragraph{Low-level controller training}
% The role of the low-level controller, $\pi^l$, is to track the given commands $a_t = [v^{cmd}_x, v^{cmd}_y, w^{cmd}_{yaw}] \in se(2)$, which consist of desired base linear velocities in the forward and lateral directions, as well as the desired yaw rate of the robot's body. 
The low-level controller, denoted as $\pi^l : \mathcal{N \times A} \xrightarrow{} \mathcal{T}$, predicts the joint position target $\boldsymbol{q}^{des}_t \sim \pi^l(\boldsymbol{q}^{des}_t|n_t, {a}_t) \in \mathbb{R}^{12}$, and converts it to robotic torque command $\boldsymbol{\tau}$ through the joint impedance controller. The low-level controller internally has an explicit estimator that predicts the body velocity $v \in \mathbb{R}^3$, feet height $z_{feet} \in \mathbb{R}^4$, and body height $z_{body} \in \mathbb{R}^1$. The learning architecture and rewards are the same as those in \cite{choi2023learning}. The low-level controller's role is to track the given commands $a_t = [v^{cmd}_x, v^{cmd}_y, w^{cmd}_{yaw}] \in se(2)$, which consist of desired base linear velocities in the forward and lateral directions, as well as the desired yaw rate. During the training of the low-level controller, velocity commands $a_t$ are randomly sampled in every iteration. \\

\paragraph{High-level controller training} 

The high-level controller consists of two distinct elements: the high-level policy, $\pi^h_{\theta}: \mathcal{L} \times \mathcal{O} \times \mathcal{A} \xrightarrow{} \mathcal{A}$, and the encoder $h_{\psi}: \mathcal{X} \xrightarrow{} \mathcal{L}$ (in simulation) or $h_{\phi}: \mathcal{O} \times \mathcal{A} \xrightarrow{} \mathcal{L}$ (in real-world). $h_{\psi}$ takes the privileged history ${X}_t \in \mathbb{R}^{22 \times N}$, consisting of $N = 20$ privileged information history $x_t \in \mathbb{R}^{22}$, and predicts the latent vector sequence $\mathbf{l}^{1:N}_t \in \mathbb{R}^{96 \times N}$. The latest encoded latent vector $\mathbf{l}^N_t$, current observation $o_t$, previous action $a_{t-1}$ are inputs for the high-level policy $\pi^h$ to predict the action ${a}_t \in se(2)$. The predicted action $a_t$ is provided at intervals of 0.2 seconds, and the low-level controller $\pi^l$ tracks the action command $a_t$ at a frequency of 0.01 seconds. To adhere to the action limits, the actions are constrained using the softmax function, limiting $v^{cmd}_x, v^{cmd}_y$, and $w^{cmd}_{yaw}$ to the range of (-1.5, 1.5) $m/s$, (-1.5, 1.5) $m/s$, and (-1.5, 1.5) $rad/s$, respectively. During the training of the high-level controller, the pre-trained low-level controller remains frozen.
\begin{align}
    &\mathbf{l}^{1:N}_t={h}_{\psi}({X}_t) \\
    &a_t\sim \pi^h_{\theta}(a_t | \mathbf{l}^N_t, o_t, a_{t-1}) \\
    &\boldsymbol{\tau}^{robot} \sim \pi^l(\boldsymbol{\tau}^{robot}|n_t, a_t) \\
    &{X}_t=\{x_t^1, ..., x_t^{N-1}, x_t^N\}
    \label{eq:trajectory}
\end{align}
We implement both ${h}_{\psi}$ and $h_{\phi}$ as a Long Short Term Memory (LSTM), and $\pi^h_{\theta}$ as Multi-Layer Perceptrons (MLPs). We concurrently train the encoder ${h}_{\psi}$ and the high-level policy $\pi^h_{\theta}$ using end-to-end model-free reinforcement learning. The reinforcement learning objective is to maximize the expected return, defined as the sum of discounted rewards over time steps $t$:

\begin{equation}
\begin{aligned}
    &J(\pi^h, \pi^l, h_{\psi}) = \mathbb{E}_{\tau \sim {p(\tau|\pi^h, \pi^l, h_{\psi})}}\bigg[\sum^{T-1}_{t=0}\gamma^tr_t\bigg]\\
\end{aligned}
\label{eq: Hierarchical structure}
\end{equation}

\paragraph{RL Reward function}

\begin{table}[h!]
\caption{Reward Functions}
\centering
\begin{tabularx}{\columnwidth}{|*{6}{>{\centering\arraybackslash}X|}}
\hline
\multicolumn{2}{|c|}{\textbf{Reward}} & \multicolumn{4}{c|}{\textbf{Expression}} \\ \hline
\multicolumn{2}{|c|}{$r_1^i$} & \multicolumn{4}{c|}{$k_1\exp(-||p_{\text{robot}} - p_{\text{object}}||_2)$} \\ \hline
\multicolumn{2}{|c|}{$r_2^e$} & \multicolumn{4}{c|}{$k_2\log(||G_{\text{goal}} - G_{\text{object}}||_2 + 0.05)$} \\ \hline
\multicolumn{2}{|c|}{$r_3^i$} & \multicolumn{4}{c|}{$k_3\exp(-\phi_{\text{robot},x}^T v_{\text{object}})$} \\ \hline
\multicolumn{2}{|c|}{$r_4^i$} & \multicolumn{4}{c|}{$k_4\exp(-(p_{\text{goal}} - p_{\text{object}})^T v_{\text{object}})$} \\ \hline
\multicolumn{2}{|c|}{$r_5^i$} & \multicolumn{4}{c|}{$k_5\exp(-||a_t - a_{t-1}||_2)$} \\ \hline
\multicolumn{2}{|c|}{$r_6^i$} & \multicolumn{4}{c|}{$k_6\exp(-||a_t - 2a_{t-1} + a_{t-2}||_2)$} \\ \hline
\multicolumn{6}{|c|}{\textbf{Intrinsic Reward}} \\ \hline
\multicolumn{6}{|c|}{
$r^i_t = \begin{cases}
    r^i_{t-1} & \text{if $||p_{\text{goal}} - p_{\text{object}}||_2 < 0.2 \ m$} \\
    k_7(r_1^i+r_3^i+r_4^i+r_5^i+r_6^i) & \text{otherwise}
\end{cases}$} \\ \hline
\multicolumn{2}{|c|}{\textbf{Extrinsic Reward}} & \multicolumn{4}{c|}{$r^e_t = k_8r_2^e$} \\ \hline
\multicolumn{6}{|c|}{\textbf{Reward Coefficients}} \\ \hline
$k_1$ & 1 & $k_2$ & 4 & $k_3$ & 1 \\ \hline
$k_4$ & 1 & $k_5$ & 2 & $k_6$ & 3 \\ \hline
$k_7$ & 1 & $k_8$ & 1 &  &  \\ \hline
\end{tabularx}

\label{table:Reward Function}
\end{table}

\begin{figure}[ht!]
    \centering
    \includegraphics[width=\columnwidth]{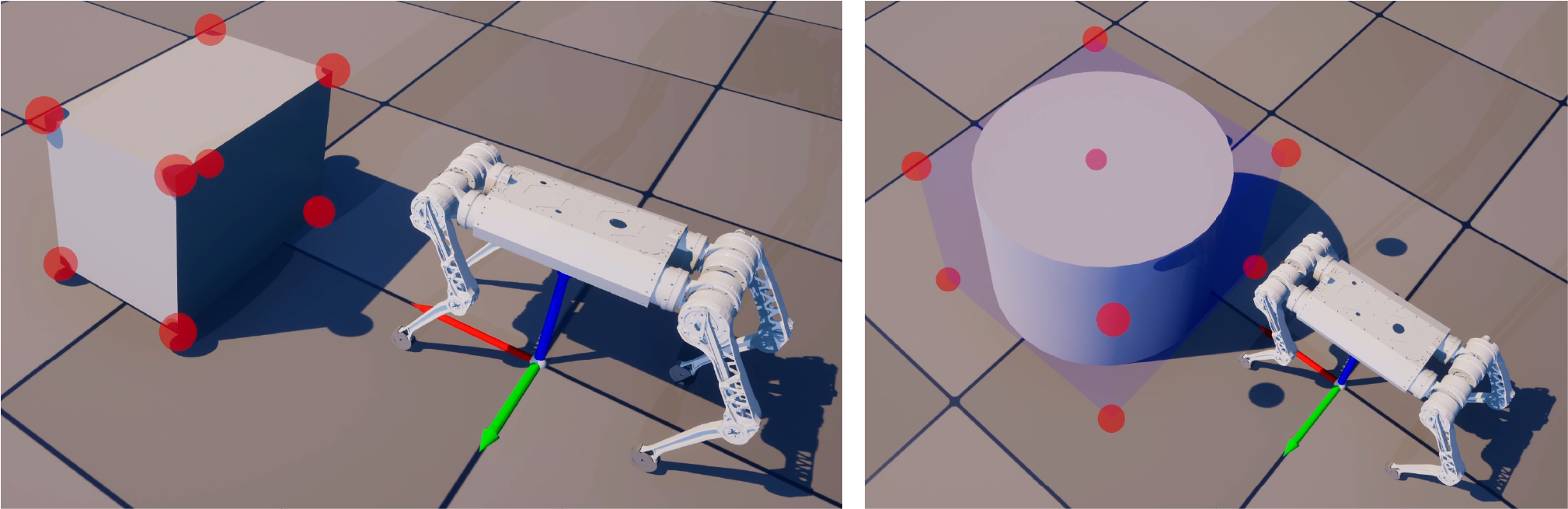}
    \caption{\textbf{8 key points}: The 8 key points are defined as rectified of the bounding box that encapsulates the object.}
\label{fig: 8 points}
\end{figure}

In the planar pushing task, the main objective is to minimize the distance between the object and the goal within the $SE(2)$ manifold. However, finding the optimal weight between rotation and position components can be challenging and may result in sub-optimal performance. To overcome this challenge, we directly minimize the distance between the key points of the object and the goal in the $\mathbb{R}^3$ space (i.e., extrinsic reward $r^e$) similar to \cite{petrenko2023dexpbt, allshire2022transferring} as shown in the Figure. \ref{fig: 8 points}. $G_{\langle \cdot \rangle} \in \mathbb{R}^{24}$ represents the set of 8 key points position, all in the robot's base frame. 

However, if we only consider extrinsic reward, the agent gets stuck in local minima when it is not in contact with the object (e.g. during the approaching phase) during which extrinsic rewards do not change. To address this, we introduce intrinsic rewards $r^i$ to encourage exploration to approach the target object. The intrinsic rewards consist of the following components: reducing the distance between the robot and the object ($r^i_1$), encouraging the object to be pushed in the robot's forward direction ($r^i_3$), inducing the object's movement direction towards the goal direction ($r^i_4$), and action smoothness rewards for smooth movements ($r^i_5, r^i_6$). The intrinsic reward is no longer updated (i.e. intrinsic reward maintains its previous value) once the robot is within a certain distance (0.2 $m$) of the object in order to prevent interrupting the extrinsic rewards to converge to the maximum. The total reward, which combines intrinsic and extrinsic rewards, is defined as $r^{total}_{t} = r^i_t + r^e_t$. \\

% \paragraph{Explicit Estimator}

% We developed an explicit estimator, denoted as $\mu: \mathcal{L} \xrightarrow{} \mathcal{X}$, with the objective of validating the incorporation of physical information about objects and environments in push behavior analysis. By taking into account the entire series of push behaviors, denoted as $\mathbf{l}^{1:N}_t$, this estimator directly predicts the privileged information $x_t \in \mathbb{R}^{22}$. This privileged information encompasses vital physical properties of the object, such as mass, inertia, and center of mass (COM), as well as environmental variables like friction, damping, and contact.

% The training process of the explicit estimator relies on supervised learning, minimizing the mean squared error (MSE) between the predictions $\mu(\mathbf{\l}^{1:N}_t)$ and the corresponding ground truth values $x^{1:N}_t$.

% \begin{equation}
% \begin{aligned}
%     L_{estimator} = ||\mu(\mathbf{l}^{1:N}) - x^{1:N}||_2
% \end{aligned}
% \label{eq:Estimator}
% \end{equation}

% Notably, our focus during training is specifically on instances after contact occurs. This choice allows us to emphasize and leverage the relevant physical information crucial for the prediction process.

\paragraph{Transferring to the real world}

The system cannot be directly deployed on the real robot because the privileged information $x_t$ is not observable. To address this limitation, we use ROA (Regularized Online Adaptation) \cite{fu2023Deep} for 1-step privileged learning. This method trains a teacher $h_{\psi}: \mathcal{X} \xrightarrow{} \mathcal{L}$ and a student $h_{\phi}: \mathcal{H} \xrightarrow{} \mathcal{L}$ simultaneously in online. The teacher $\psi$ takes the privileged information history $X = \{x^1, ..., x^{N-1}, x^N\}$ as input and predicts a latent vector $\mathbf{l}$ to implicitly encode object properties and adapt the policy to various environments. Meanwhile, the student $h_{\phi}$ imitates the teacher $h_{\psi}$ to approximate $\tilde{\mathbf{l}}$ based on the observation and action history $H = \{o^1, a^1, ..., o^{N-1}, a^{N-1}, o^N, a^N\}$. This enables the student to estimate the encoded object properties from observable data. Here, the stacking process is asynchronous: the observation and privileged information history is updated at the low-level policy frequency (100Hz), and the action history is updated at the high-level policy frequency (5Hz). This is because state transitions occur after every low-level policy execution. The teacher $\psi$ and the student $\phi$ are trained simultaneously using reinforcement learning \eqref{eq: Hierarchical structure} with the following auxiliary loss:

\begin{equation}
\begin{aligned}
    L_{adaptation} = \lambda||\tilde{\mathbf{l}} - sg[\mathbf{l}]||_2 + ||sg[\tilde{\mathbf{l}}] - \mathbf{l}||_2
\end{aligned}
\label{eq:adaptive}
\end{equation}

where $sg[\cdot]$ is the stop gradient operator, and $\lambda$ is the Lagrangian multiplier. The first term is for supervision, and the second term is for regularization. The entire high-level controller training procedure is presented in \ref{algo: entire process}.\\

\begin{algorithm}[!ht]
\DontPrintSemicolon
  \For{episode=1, M} {
    {Initialize $h_\psi$, $h_\phi$} \\
    {Initialize action history $A_{queue}$ to capacity N} \\
    {Initialize state history $S_{queue}$ to capacity N} \\
    {Initialize data batch $D$} \\
    \tcc{High-level controller, 5Hz}
    \For{$t = 1,T$} 
    {
        $H = \langle O_{history}, A_{queue} \rangle$ \tcp*{$O_{history} \in S_{queue}$}
        $\mathbf{l}^{1:N}_t$ = $h_\psi(X_{history})$ \tcp*{$X_{history} \in S_{queue}$}
        $\tilde{\mathbf{l}}^{1:N}_t$ = $h_\phi(H)$ \tcp*{$\tilde{\tau} \in \tau$}
        $a_t \sim \pi_{\theta}^h(a_t | l^N_t, o_t, a_{t-1})$ \\
        $A_{queue} \xleftarrow[]{} a_t$ \\
        \tcc{Low-level controller, 100Hz}
        \For{${t' = 1,K}$} 
        {
            Observe state $s_{t'}$, and stack $S_{queue} \xleftarrow[]{} s_{t'}$ \\
            $\boldsymbol{\tau}^{robot}_{t'} = \pi^l_{\kappa}(n_{t'}, a_t)$ \tcp*{$n_{t'} \in s_{t'}$} 
            Execute $\boldsymbol{\tau}^{robot}_{t'}$ \\
        }
        Store transition $s_{t'}, a_t, r_{t'}, \mathbf{l}_t, \tilde{\mathbf{l}}_t$ in $D$
    }
    \tcc{$(\mathbf{l}, \tilde{\mathbf{l}}, r, s) \sim D$}
    $L_{adaptation} = \lambda||\tilde{\mathbf{l}} - sg[\mathbf{l}]||_2 + ||sg[\tilde{\mathbf{l}}] - \mathbf{l}||_2$ \\
    % $L_{estimator} = ||\mu_{\zeta}(\mathbf{l}^{1:N}) - x^{1:N}||_2$ \tcp*{sample after contact occurs}
    $J = \mathbb{E}_{\tau \sim {D}}\big[\sum^{T}_{t=1}\gamma^tr_{t'}\big]$ \\
    $L_{total} = -J + L_{adaptation}$ \\
    Compute $\nabla_{\theta,\psi, \phi}L_{total}$ and update \tcp*{PPO \cite{schulman2017proximal}}
  }

\caption{Learning the high-level controller in simulation (with fixed low-level controller)}
\label{algo: entire process}
\end{algorithm}

% \subsection{Training Details} \label{Trainign Details}

\paragraph{Training Environment Generation}

\begin{table}[ht!]
\caption{Training and testing parameters in simulation}
\centering
\resizebox{\columnwidth}{!} {
\begin{tabular}[t]{|l|c|c|c|}
\hline
\multicolumn{2}{|c|}{\textbf{Parameters}} & \textbf{Training Range} & \textbf{Testing Range} \\ \hline
\multicolumn{2}{|c|}{Mass ($kg$)} & $U$(6, 18) & $U$(10, 20) \\ \hline
\multicolumn{2}{|c|}{Inertia axis angle $\theta$ \ (degree)} & $U$(-20, 20) & $U$(-20, 20)  \\ \hline
\multicolumn{2}{|c|}{COM volume $V$ ($\%$)} & $U$(0,50) & $U$(0,50) \\ \hline
\multicolumn{2}{|c|}{Friction} & $U$(0.2,0.4) & $U$(0.15, 0.4) \\ \hline
\multicolumn{2}{|c|}{Drag coefficient} & $U$(0.3,0.7) & $U$(0.0, 1.0) \\ \hline
\multicolumn{2}{|c|}{Object type} & [Box, Cylinder] & [Box, Cylinder] \\ \hline
\multirow{4}{*}{Object dimension} 
& Diameter ($m$) & $U$(0.5, 1.5) & $U$(0.5, 1.5) \\
& Width ($m$) & $U$(0.5, 1.5) & $U$(0.5,1.5) \\ 
& Depth ($m$) & $U$(0.5, 1.5) & $U$(0.5, 1.5) \\
& Height ($m$) & $U$(0.35, 1.05) & $U$(0.35, 1.05) \\ \hline
\end{tabular}
\label{table:environment parameters}
}
\end{table}

During the training high-level controller, we utilized 300 parallel environments in simulation to efficiently collect data. Each environment was equipped with a $\textbf{Random Object Generator}$ that spawns objects with various shapes and physical parameters. Table \ref{table:environment parameters} shows the training and testing ranges for the object's physical parameters. All variables are sampled from uniform distributions denoted as $U$. The parameters are randomly sampled in every episode. The center of mass (COM) is randomly chosen within $V = 50\%$ of the object's volume, starting from the geometric center of mass $(x_0, y_0, z_0)$. To introduce diversity in the object's inertia, we diagonalize the original inertia tensor $I$ to obtain the principal moments of inertia $(I_1, I_2, I_3)$ and principal axes $(u_1, u_2, u_3)$. We then transform the principal axis by an angle $\theta$. The objects generated in the simulation include cylinders and boxes, each represented as a one-hot vector to indicate their shape. The object dimension is represented as a real-valued vector with three components: for cylinders $\{$diameter, diameter, height$\}$, and $\{$width, depth, height$\}$ for boxes. All the physical parameters, shape, and dimension information of the objects belong to the privileged information $x_t$.
% This approach allows the robot to learn how to manipulate different objects and adapt in real-time, even when the physical properties of the objects are suddenly changed.

\section{Experiment and result}
\label{sec:result}

We evaluate the capabilities of our framework both in simulation and the real world. We measure the success rate in the following task: re-posing objects using a quadruped robot under various conditions, including different object shapes and inertial parameters. We also examine the effectiveness of our system by comparing its performance against several baseline methods in simulation.

The training process was conducted in Raisim \cite{raisim} using an AMD Ryzen9 5950X CPU and an NVIDIA GeForce RTX 3070 GPU. The training involved 300 environments and 10,000 iterations, with each iteration taking approximately 20 seconds in simulation. The overall training process took around 15 hours. For the deployment on the real robot, the trained networks were implemented on an Intel NUC (4-core) PC embedded in the robot. The pose information of the object and the robot was obtained in real-time using the VICON vero v2.2 motion capture system shown in Figure. \ref{fig: Vicon}.

\subsection{Simulation Result}

In the simulation experiment, we randomly generate 1000 different situations based on the testing parameters in Table \ref{table:environment parameters}. We evaluate the performance of the system using five criteria, each with different tolerance distances ($d$) and angles ($\theta$) for the desired pose, as detailed in Table \ref{table:baseline}. Both the objects and the robot are spawned randomly with each position and orientation within 4 $m$ and between 0 to 2$\pi$ with respect to the world frame. The desired pose for the object is sampled within a range of translation up to 4 $m$ and orientation from 0 to 2$\pi$ from the object's initial pose. A trial is considered successful if the robot manipulated the object to the desired pose for each criterion within 30 seconds. We assess the success rate and the average time taken by the successful trials. We compare the performance of our method with the following baselines:

\begin{itemize}
    \item w/o Pre-trained controller: In order to validate the performance enhancement of our hierarchical control structure, we exclude a pre-trained low-level controller, and instead high-level controller outputs joint position target $q^{des}_t$ directly.
    \item w/o Adaptation (Domain Randomization): We train the high-level controller without an encoder, and the encoded latent vector input $\mathbf{l}^N$ was removed. Instead, we randomize the physical properties of the objects during training.
    \item w/o 8 key points: To validate the effectiveness of the 8 key points, we replace the extrinsic reward with the sum of the L2 norms of position and orientation errors.
    \item MLP Encoder: To verify whether the recurrent structure better captures the physical properties information of objects and the environment, we replace the LSTM encoder with an MLP encoder.
    \item w/o Inertial parameters: To verify whether our latent embedding actually estimates object inertial parameters, we exclude the object inertial parameters, including mass, inertia, and center of mass (COM) from the privileged information $x_t$.
    \item w/o Intrinsic switch: To verify whether considering both extrinsic and intrinsic rewards, even when close to the goal, can lead to sub-optimal behavior, we trained without intrinsic switching.
    \item Expert: We use the true value of the privileged information $x_t$, and the encoder $h_{\psi}$ in simulation.
\end{itemize}

\begin{table*}[!ht] 
\caption{Simulation Testing Results}
\resizebox{\textwidth}{!} {
\centering \begin{tabular}{ccccccccccccc} 

\toprule 
\multirow{4}{*}{\textbf{Method}} & \multicolumn{10}{c}{\textbf{Success Rate} ($\%$)} & \multirow{4}{*}{\begin{tabular}[c]{@{}c@{}}\textbf{Time}\tabularnewline (sec)\end{tabular}} & \multirow{4}{*}{\textbf{Observable}} \tabularnewline\cmidrule(l){2-11}
& \multicolumn{2}{c}{Criteria 1} & \multicolumn{2}{c}{Criteria 2} & \multicolumn{2}{c}{Criteria 3} & \multicolumn{2}{c}{Criteria 4} & \multicolumn{2}{c}{Criteria 5} &  \tabularnewline\cmidrule(l){2-3}\cmidrule(l){4-5}\cmidrule(l){6-7}\cmidrule(l){8-9}\cmidrule(l){10-11} & \multicolumn{1}{c}{$d=0.05 m$} & \multicolumn{1}{c}{$\theta=5^\circ$} & \multicolumn{1}{c}{$d=0.05 m$} & \multicolumn{1}{c}{$\theta=10^\circ$} & \multicolumn{1}{c}{$d=0.05 m$} & \multicolumn{1}{c}{$\theta=15^\circ$} & \multicolumn{1}{c}{$d=0.1 m$} & \multicolumn{1}{c}{$\theta=10^\circ$} & \multicolumn{1}{c}{$d=0.03 m$} & \multicolumn{1}{c}{$\theta=5^\circ$} & &    \tabularnewline \midrule 
w/o Pre-trained controller & \multicolumn{2}{c}{0} & \multicolumn{2}{c}{0}  & \multicolumn{2}{c}{0} & \multicolumn{2}{c}{0}  & \multicolumn{2}{c}{0} & - & \checkmark \tabularnewline 
w/o Adaptation  & \multicolumn{2}{c}{58.4} & \multicolumn{2}{c}{56.1} & \multicolumn{2}{c}{60.9} & \multicolumn{2}{c}{67.4} & \multicolumn{2}{c}{49.8} & 18.3 & \checkmark \tabularnewline
MLP Encoder  & \multicolumn{2}{c}{79.5} & \multicolumn{2}{c}{76.3} & \multicolumn{2}{c}{80.4} & \multicolumn{2}{c}{91.3} & \multicolumn{2}{c}{56.6} & 17.2 & \checkmark \tabularnewline 
w/o 8 key points & \multicolumn{2}{c}{60.5} & \multicolumn{2}{c}{47.0}& \multicolumn{2}{c}{82.5} & \multicolumn{2}{c}{61.0}  & \multicolumn{2}{c}{45.0}  & 14.5 & \checkmark  \tabularnewline 
w/o Intrinsic switch & \multicolumn{2}{c}{91.6} & \multicolumn{2}{c}{91.0}& \multicolumn{2}{c}{91.6} & \multicolumn{2}{c}{92.9}  & \multicolumn{2}{c}{82.5}  & 13.2 & \checkmark  \tabularnewline 
w/o Inertial parameters & \multicolumn{2}{c}{84.5} & \multicolumn{2}{c}{82.6}& \multicolumn{2}{c}{85.1} & \multicolumn{2}{c}{89.3}  & \multicolumn{2}{c}{68.2}  & 17.3 & \checkmark  \tabularnewline 
\textbf{Ours} & \multicolumn{2}{c}{\textbf{96.1}} & \multicolumn{2}{c}{\textbf{96.1}}& \multicolumn{2}{c}{\textbf{96.3}} & \multicolumn{2}{c}{\textbf{96.8}}  & \multicolumn{2}{c}{\textbf{93.6}} & \textbf{11.23} & \checkmark \tabularnewline \midrule 
Expert  & \multicolumn{2}{c}{97.8} & \multicolumn{2}{c}{97.8}& \multicolumn{2}{c}{97.9} & \multicolumn{2}{c}{98.0}  & \multicolumn{2}{c}{96.7}  & 10.5 & -   \tabularnewline\bottomrule

\end{tabular}
\label{table:baseline}
} 

\end{table*}

Table \ref{table:baseline} shows our simulation result. The proposed model shows slightly lower accuracy compared to the expert. The MLP encoder $h_{\langle \cdot \rangle}$ resulted in a loss of performance, indicating the difficulty in determining push behavior based solely on momentary data history. In the case without adaptation, the high-level policy demonstrates difficulties in decision-making, highlighting the importance of considering object physical properties inherently. When the pre-trained controller is not used, the robot failed to stand up or locomote properly, resulting in failure in all test cases. Additionally, without using the 8 key points, the performance significantly degraded, especially in terms of poor orientation alignment, as noted in criteria 1 to 3. Furthermore, the absence of an intrinsic switch led to decreased performance, particularly in criterion 5. This indicates that considering intrinsic rewards after a certain point can actually interfere with reaching the optimal behavior. Without considering inertial parameters, criterion 5 shows a severe performance loss, suggesting that the delicate manipulation of objects requires the consideration of their inertial parameters. This demonstrates that the proposed system effectively captures object inertial parameters via encoded latent vectors. Overall, our method outperformed the baseline methods, demonstrating its effectiveness in learning pushing behavior for a robot in a simulated environment.

\subsection{Real-world Experiment Result}

\begin{figure}[ht!]
    \centering
    \begin{subfigure}[b]{0.49\columnwidth}
        \centering
        \includegraphics[height=0.14\textheight]{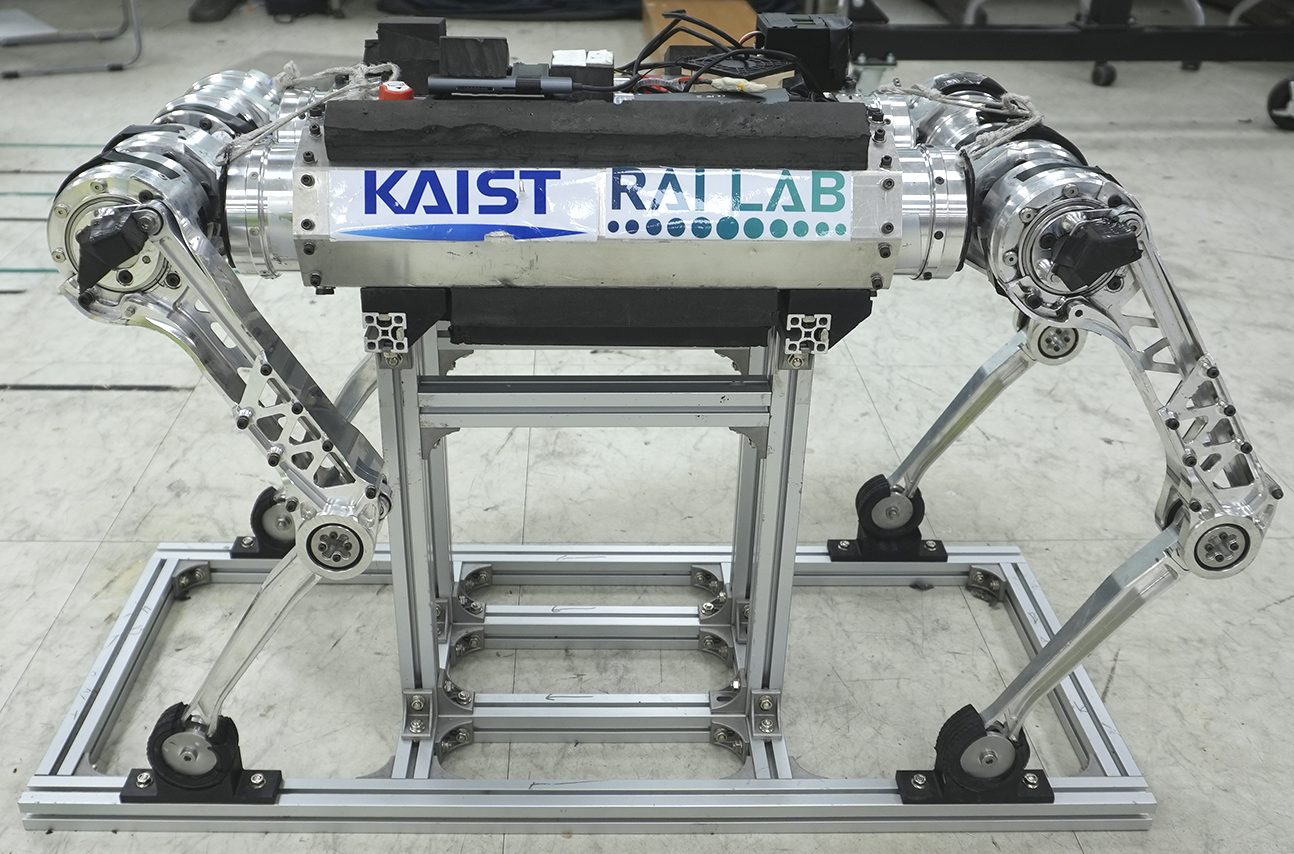}
        \caption{Robot platform}
        \label{fig: Raibo}
    \end{subfigure}
    \begin{subfigure}[b]{0.49\columnwidth}
        \centering
        \includegraphics[height=0.14\textheight]{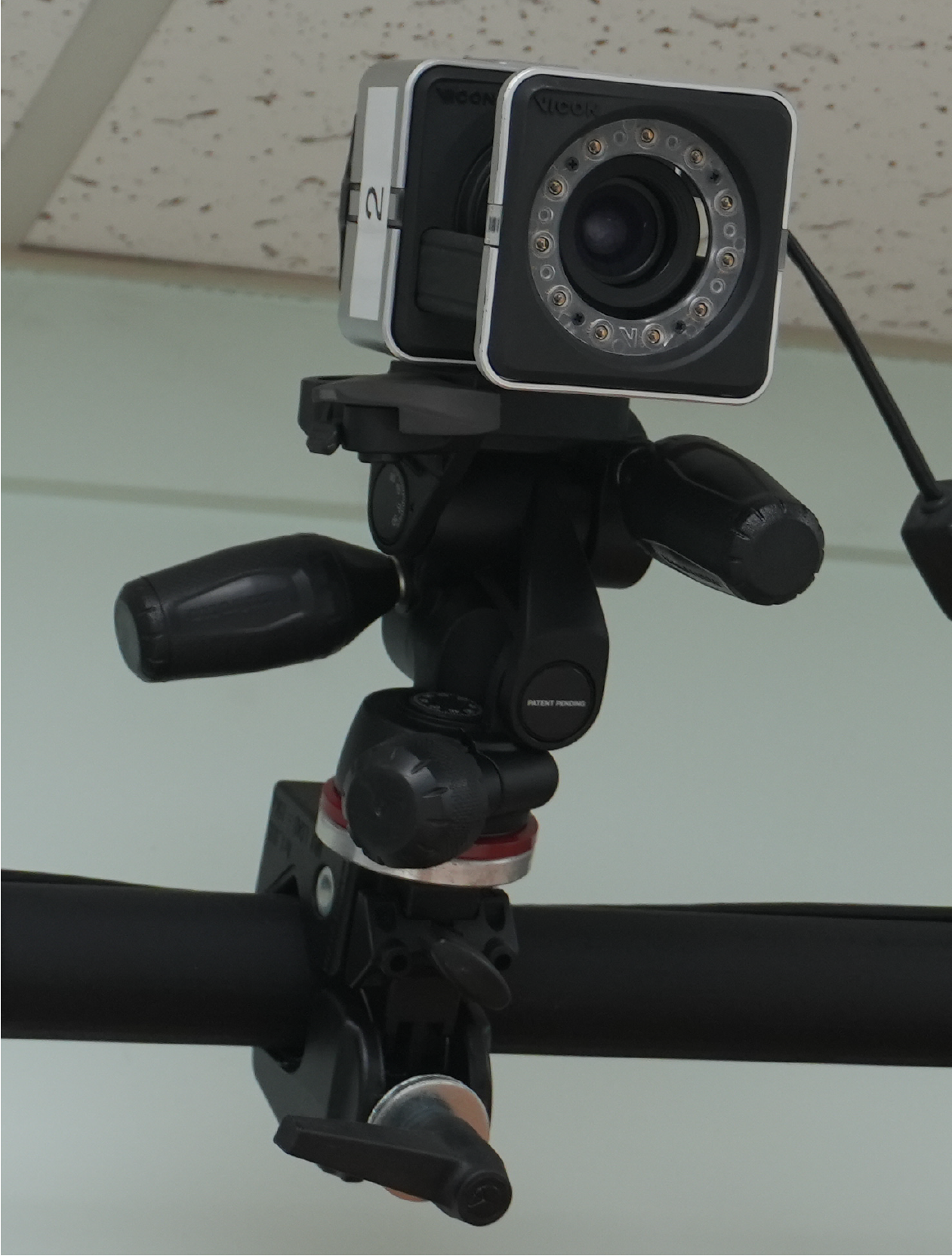}
        \caption{Motion capture system}
        \label{fig: Vicon}
    \end{subfigure}
    \caption{\textbf{Real-world system}: (a) Raibo weights 27 $kg$ and has a torque limit of 55 $Nm$, designed and built in-house. (b) VICON provides pose information for objects and robot at a frequency of 200Hz.}
\end{figure}

\begin{table}[!ht]
\caption{Real-world experiment object properties}
\centering
\resizebox{\columnwidth}{!} {
\begin{tabular}[t]{|l|c|c|c|}
\hline
\textbf{Shape} & \textbf{Category} & \textbf{Mass} ($kg$) & \textbf{Dimension} ($m$) \\
\hline
\multirow{3}{*}{Box} & Yellow & 15.3 & (Width,Depth,Height) = (0.45, 0.7, 0.5)\\
  &  Green & 14.2 & (Width,Depth,Height) = (0.4, 0.6, 0.45) \\
  &  Black & 12.6 & (Width,Depth,Height) = (0.3, 0.55, 0.65) \\
\hline
\multirow{2}{*}{Cylinder} & Trash bin & 17.4 & (Diameter,Height) = (0.56, 0.7) \\
& Drum & 19.2 & (Diameter,Height) = (0.4, 0.55) \\
\hline
\end{tabular}
}
\label{table: real-world object}
\end{table}

\begin{figure}[ht!]
    \centering
    \includegraphics[width=1.0\columnwidth]{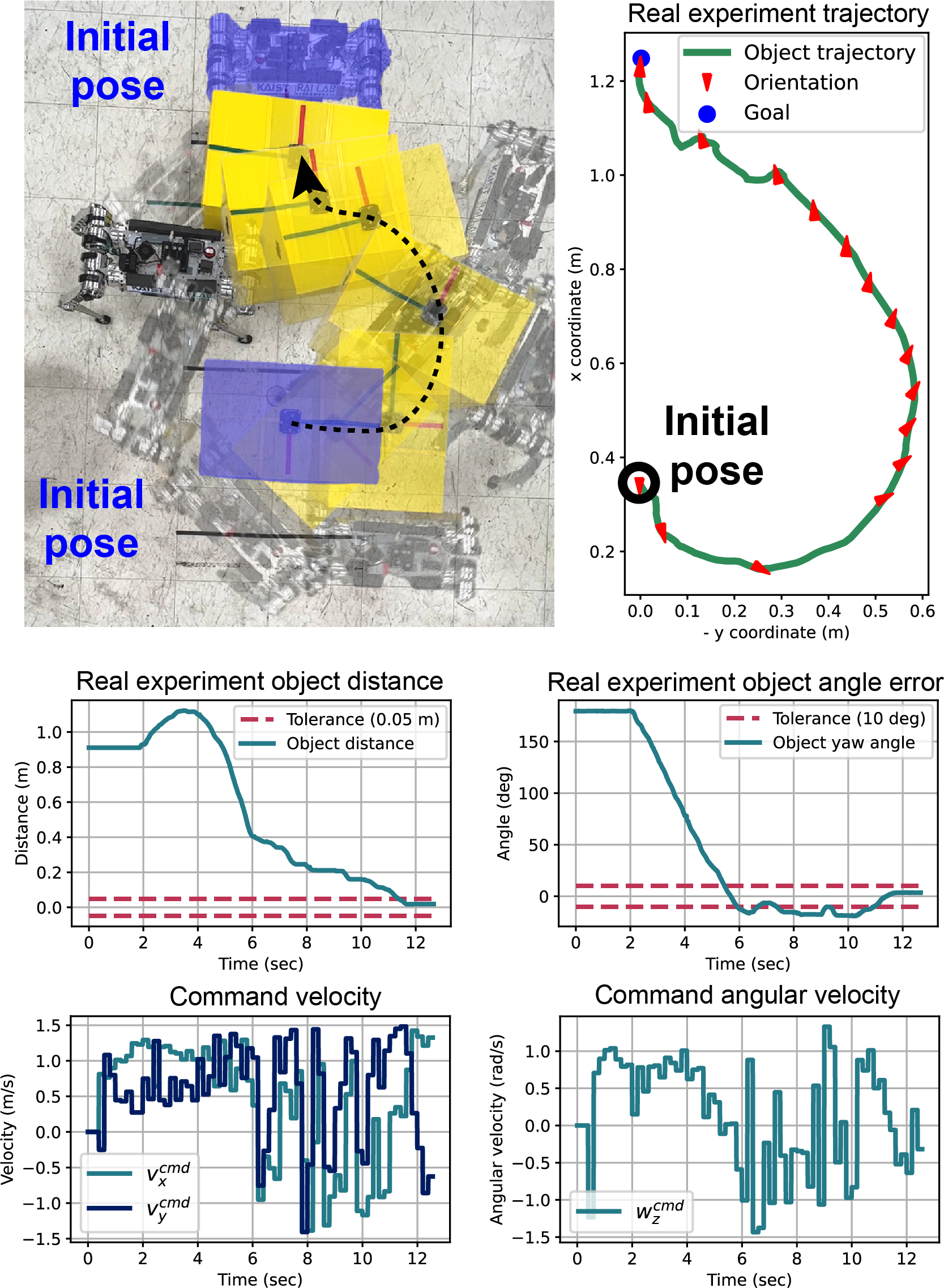}
    \caption{\textbf{Real-world experiment result:} The figure shows the trajectory of the object and the $SE(2)$ error with respect to the goal pose. The command velocity predicted by the high-level controller during the experiment is shown below.}
    \label{fig: result trajectory}
\end{figure}

For real-world experiments, we use the Raibo (Figure. \ref{fig: Raibo}). We demonstrate the performance of our system on box and cylindrical shaped objects with various physical properties, including time-varying ones. Since the motion capture system area is limited, we fix the desired pose and the object's position and vary the object's orientation and the robot's pose. Three experiments were conducted for each object, with each initial orientation being 45$^\circ$, 90$^\circ$, or 180$^\circ$. We define success as an object's final pose being within a certain tolerance, 0.05 $m$ and 10$^\circ$. The object properties used in the experiment are summarized in Table \ref{table: real-world object}. The objects were packed with furniture or water to change the center of mass and inertia in real-time during manipulation. In a total of 45 trials, (with 5 objects, 3 initial orientations, and 3 trials for each case) our system achieved without a single failure for all testing cases even though the robot had no prior knowledge of the object's physical properties. We directly deployed our system in the real world without any fine-tuning (zero-shot learning). We present one of the real-world experiment results in Figure. \ref{fig: result trajectory}. 

% This was possible because our system was trained on a diverse set of objects with varying physical properties via Random Object Generator, which allowed it to learn the underlying principles of object manipulation. We also used an encoder to implicitly adapt to the physical properties of the objects during manipulation.

\section{Conclusion and Future Works}
We have proposed a learning-based whole-body manipulation approach with a hierarchical structure for planar-pushing tasks. The proposed method allows the robot to manipulate objects without prior knowledge of their physical properties. Our system is computationally efficient, as it can perform both approaching and pushing with NN predictions, without the need for trajectory optimization or contact reasoning. We have demonstrated the effectiveness of our approach through numerical and experimental validations. In simulations, we have shown that the proposed method achieves a remarkable success rate under diverse object physical properties. The system also demonstrates high accuracy in real-world robot experiments, manipulating objects up to 70 $\%$ of the robot's weight. We believe that this study could point to numerous directions for future research in developing more general methods for whole-body manipulation and loco-manipulation. Additionally, we are currently investigating methods for replacing the motion capture system with exteroceptive sensor data to avoid being restricted to a specific workspace. We are also interested in obstacle avoidance in cases where there are obstacles in the robot's path.

\bibliographystyle{IEEEtran}
\bibliography{IEEEabrv,reference}

\end{document}